\documentclass[11pt,a4paper]{article}
\usepackage[hyperref]{acl2020}
\usepackage{times}
\usepackage{latexsym}

\usepackage{graphicx}

\usepackage{microtype}

\usepackage{amsmath}
\usepackage{amsfonts}
\usepackage[noend]{algpseudocode}
\usepackage{algorithmicx,algorithm}
\usepackage{booktabs}
\usepackage{multirow}
\usepackage{color,xcolor}
\usepackage{soul}

\aclfinalcopy 
\def\aclpaperid{609} 

\title{Generate, Delete and Rewrite: A Three-Stage Framework for Improving Persona Consistency of Dialogue Generation}

\author{Haoyu Song$^{1}$\thanks{\ \ This work was done when Haoyu Song was an intern at Tencent AI Lab.},\ \ Yan Wang$^2$,\ \ Wei-Nan Zhang$^{1}$,\ \ Xiaojiang Liu$^2$,\ \ Ting Liu$^1$ \\
  $^1$Research Center for Social Computing and Information Retrieval\\
  Harbin Institute of Technology, Heilongjiang, China\\
  $^2$ Tencent AI Lab, Shenzhen, China\\
  \texttt{\{hysong,wnzhang,tliu\}@ir.hit.edu.cn}\\
  \texttt{\{brandenwang,kieranliu\}@tencent.com}\\
  }

\date{}

\begin{document}
\maketitle
\begin{abstract}

Maintaining a consistent personality in conversations is quite natural for human beings, but is still a non-trivial task for machines. 
The persona-based dialogue generation task is thus introduced to tackle the personality-inconsistent problem by incorporating explicit persona text into dialogue generation models.
Despite the success of existing persona-based models on generating human-like responses, their one-stage decoding framework can hardly avoid the generation of inconsistent persona words.
In this work, we introduce a three-stage framework that employs a generate-delete-rewrite mechanism to delete inconsistent words from a generated response prototype and further rewrite it to a personality-consistent one.
We carry out evaluations by both human and automatic metrics. Experiments on the Persona-Chat dataset show that our approach achieves good performance.

\end{abstract}

\section{Introduction}

In an open-domain conversation scenario, two speakers conduct open-ended chit-chat from the initial greetings and usually come to focus on their characteristics, such as hobbies, pets, and occupations, etc., in the course of the conversation.
For humans, they can easily carry out conversations according to their personalities~\citep{song-percvae}, but fulfilling this task is still a challenge for recent neural dialogue models~\citep{WelleckDNLI}.

\begin{figure}
\centering
\includegraphics[width=.97\columnwidth]{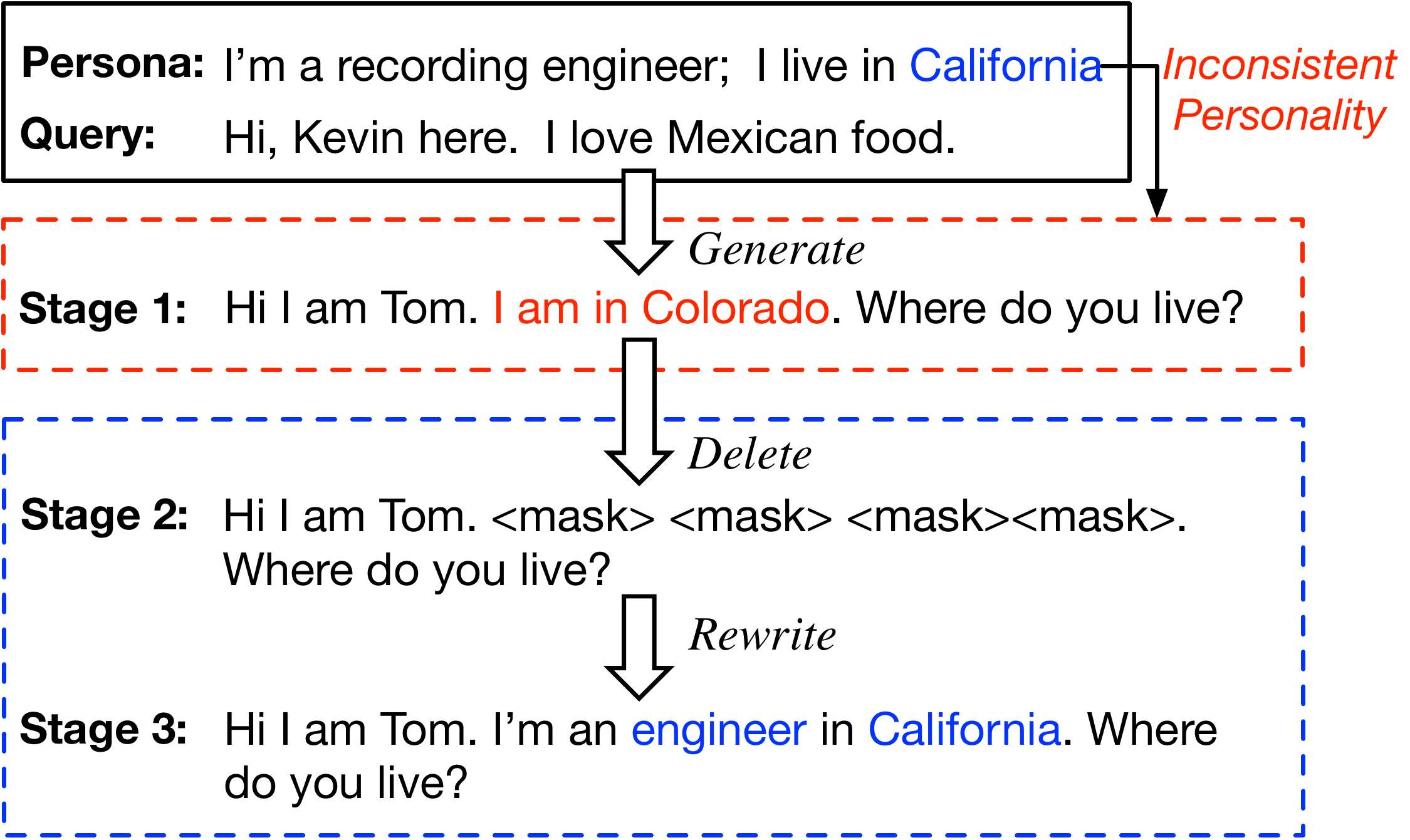}
\caption{ A common problem for persona-based dialogue models is that they can hardly avoid the generation of inconsistent persona words. Although the model generates a response which looks good, it is an inconsistent one. With further rewriting, the model can focus more on improving persona consistency. }
\label{fig:1}
\end{figure}

One main issue is that these models are typically trained over millions of dialogues from different speakers, and the neural dialogue models have a propensity to mimic the response with the maximum likelihood in the training corpus~\citep{li-etal-2016-persona}, which results in the frequent inconsistency in responses~\citep{zhang-2018-persona}.
Another issue is the user-sparsity problem~\citep{qian2017assigning} in conventional dialogue corpora~\citep{serban2015survey}. Some users have very few dialogue data, which makes it difficult for neural models to learn meaningful user representations~\citep{li-etal-2016-persona}.

To alleviate the above issues,~\citet{zhang-2018-persona} introduced the Persona-Chat dataset to build more consistent dialogue models. Different from conventional dialogue corpora, this dataset endows dialogue models with predefined personas, which is in the form of textually described profile (as shown in the first line of Figure~\ref{fig:1}). The persona-based dialogue models also adopt an encoder-decoder architecture and are enhanced with persona encoding components, such as memory network~\citep{memorynetworks} and latent variable~\citep{kingma2013vae}. These models turn out to produce more consistent responses than the persona-free ones~\citep{zhang-2018-persona,song-percvae}.

Despite the successful application of the encoder-decoder framework in persona-based dialogue models, one concern is that they lack extra attention to the key persona information. The model will learn to minimize the overall loss of every decoded word, but this may lead to the neglect of the key personas: change of one persona-related word may not significantly affect the overall loss, but could turn a good response into a totally inconsistent one.
As shown in Stage 1 of Figure~\ref{fig:1}, only one improper word ``Colorado'' leads the response to be inconsistent.

A desirable solution should be able to capture personas and automatically learn to avoid and refine inconsistent words before the response.
In this paper, we present a Generate-Delete-Rewrite framework, GDR, to mitigate the generation of inconsistent personas. 
We design three stages specifically for the goal of generating persona consistent dialogues:
The first {\it Generate} stage adopts a transformer-based generator to produce a persona-based response prototype. 
The second {\it Delete} stage employs a consistency matching model to identify inconsistencies and delete (by masking) the inconsistent words from the prototype.
Finally, in the {\it Rewrite} stage, a rewriter polishes the masked prototype to be more persona consistent.
To examine the effectiveness of our GDR model, we carried out experiments on the public available Persona-Chat dataset~\citep{zhang-2018-persona}.

We summarize the main contributions as follows:
\begin{itemize}
  \item A three-stage end-to-end generative framework, GDR, was proposed for the generation of persona consistent dialogues.
  \item A matching model was integrated into the generation framework to detect and delete inconsistent words in response prototype.
  \item Experimental results show the proposed approach outperforms competitive baselines on both human and automatic metrics.
\end{itemize}

\section{Related Work}

End-to-end dialogue generation approaches are a class of models for building open-domain dialogue systems, which have seen growing interests in recent years~\citep{vinyals2015neural,shang2015neural,serban2016building,li_deeprl,zhao2017dialoguecvae,li2017adversarial}. These dialogue models adopted recurrent units 
in a sequence to sequence ({\it seq2seq}) fashion~\citep{sutskever2014sequence}.
Since the {\it transformer} has been shown to be on par with or superior to the recurrent units~\citep{vaswani2017attention}, some dialogue models began to take advantage of this architecture for better dialogue modeling~\citep{dinan2018wizard,su-rewriter}.

Besides the advancements in dialogue models, the emergence of new dialogue corpus has also contributed to the research field.
\citet{zhang-2018-persona} introduced the Persona-Chat dataset, with explicit persona texts to each dialogue. 
Based on seq2seq model and memory network, they further proposed a model named {\it Generative Profile Memory Network} for this dataset. Following this line,~\citet{yavuz2019deepcopy} designed the {\it DeepCopy} model, which leverages copy mechanism to incorporate persona texts.~\citet{song-percvae} integrated persona texts into the {\it Per-CVAE} model for generating diverse responses. However, the persona-based models still face the inconsistency issue~\citep{WelleckDNLI}. To model the persona consistency,~\citet{WelleckDNLI} annotated the Persona-Chat dataset and introduced the Dialogue Natural Language Inference (DNLI) dataset. This dataset converts the detection of dialogue consistency into a natural language inference task~\citep{bowman2015large}.

Personalized dialogue generation is an active research field~\citep{li-etal-2016-persona,qian2017assigning,zhang-2018-persona,zheng2019personalized,zheng2019pre,zhang2019neural}. In parallel with this work,~\citet{song2019generating} leveraged adversarial training to enhance the quality of personalized responses.~\citet{liu-etal-2020-personachat} incorporated mutual persona perception to build a more explainable~\citep{liu-etal-2019-towards-explainable} dialogue agent.
Other relevant work lies in the area of multi-stage dialogue models~\citep{lei2020estimation}. Some retrieval-guided dialogue models~\citep{weston2018retrieve,wu2019response,cai2019skeleton,cai2019retrieval,su2020prototype} also adopted a multi-stage framework, but the difference from our work is obvious: we generate the prototype rather than retrieve one. A high-quality retrieved response is not always available, especially under the persona-based setting. 

\section{Model}

\subsection{Overview}
In this work, we consider learning a generative dialogue model to ground the response with explicit persona. We focus on the persona consistency of single-turn responses, and we leave the modeling of multi-turn persona consistency as future work.

\begin{figure*}[ht]
\centering
\includegraphics[width=0.985\linewidth]{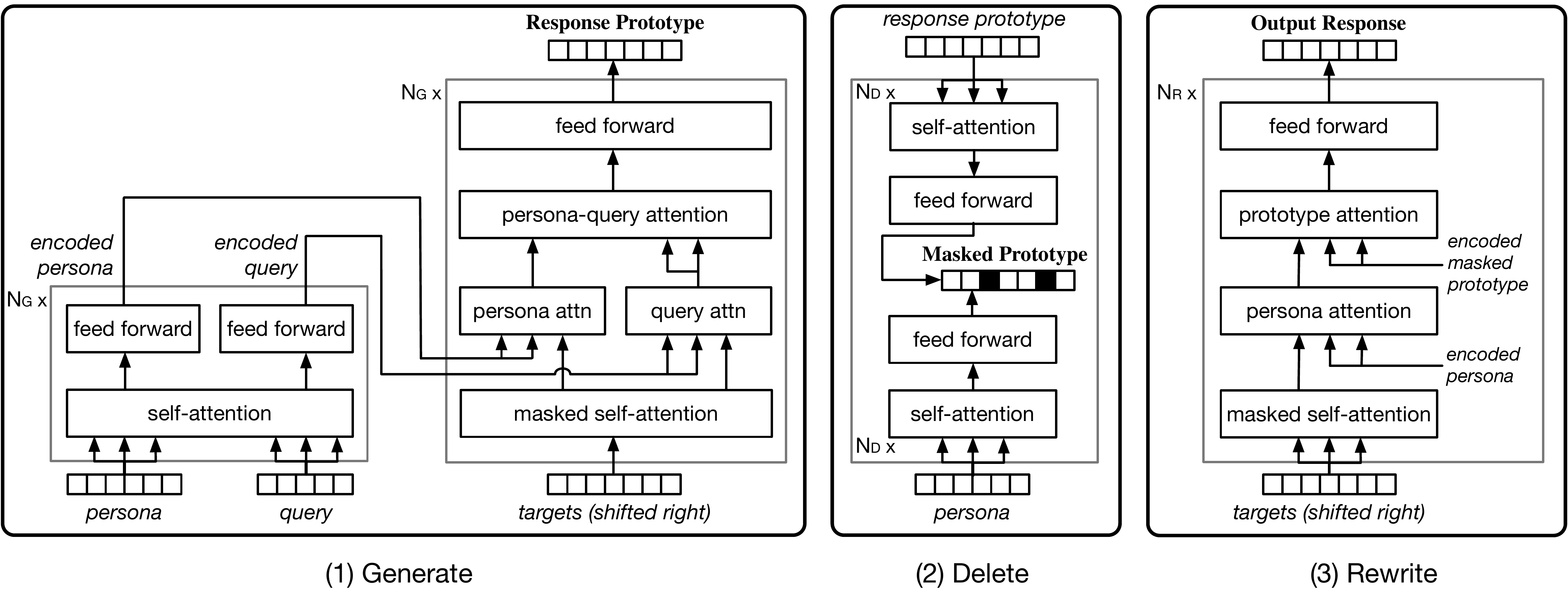}
\caption{ The overall architecture of our three-stage GDR model, including a prototype generator (Generate stage), a consistency matching model (Delete stage), and a masked prototype rewriter (Rewrite stage). The italics denote the inputs of each stage, and the boldfaces denote the outputs. All the attentions (attn) here refer to the multi-head attention. For the sake of brevity, we omitted some layers of the transformer in this figure.
}
\label{fig:overall_model}
\end{figure*}

Formally, we use uppercase letters to represent sentences and lowercase letters to represent words.
Let $Q=q_1,q_2,...,q_n$ denotes the input query with $n$ words, and let $P=\{P^{(1)},P^{(2)},...,P^{(k)}\}$ be the $k$ different persona texts, where $P^{(i)}=p^{(i)}_1,p^{(i)}_2,...,p^{(i)}_{m_i}$ is the $i$-th persona text with ${m_i}$ words. Our goal is to learn a dialogue model $\mathbb M$ to generate a response $\hat Y=y_1,y_2,...,y_k$, which is consistent with the persona, based on both query $Q$ and persona $P$. In abbreviation, $\hat Y={\mathbb M}(Q,P)$.

More concretely, as shown in Figure~\ref{fig:overall_model}, the proposed model $\mathbb M$ consists of three parts: 

1) Prototype generator {$\text G$}. This component takes persona texts and query as input and generates a response prototype for further editing.
It adopts an encoder-decoder architecture~\citep{sutskever2014sequence}, with the transformer~\citep{vaswani2017attention} applied in both the encoder and the decoder.

2) Consistency matching model {$\text D$}. This model is designed to detect and delete those words in the prototype that could lead to inconsistency. We train this model in a natural language inference fashion on the DNLI~\citep{WelleckDNLI} dataset. 

3) Masked prototype rewriter {$\text R$}. The rewriter learns to rewrite the response prototype to a more consistent one.
It is also a transformer decoder, which adopts a similar architecture as the decoder of {$\text G$}. The difference lies in that it takes the masked prototype, rather than the query, as input. 

\subsection{Generate: Prototype Generator}
\label{sec:GTG}

We apply the encoder-decoder structure to build our prototype generator $\text G$. For the encoder, we use the self-attentive encoder in the transformer. For the decoder, built upon the transformer decoder, we propose a tuple-interaction mechanism to model the relations among persona, query, and response.

\subsubsection*{Self-Attentive Encoder}
\label{sec:SAE}
As the persona $P$ is composed of several sentences, we unfold all words in $P$ into a sequence $p^{(1)}_1,p^{(1)}_2,...,p^{(i)}_{m_j},...,p^{(k)}_{m_k}$. 

Then we use the self-attentive encoder~\citep{vaswani2017attention} to compute the representations of the persona texts and query separately. 
The multi-head attention is defined as $\text{MultiHead}(Q,K,V)$, where $Q$,$K$,$V$ are query, key, and value, respectively.
The encoder is composed of a stack of $N_G$ identical layers. 
Take the first stack encoding of $P$ for example:
\begin{gather}
\label{formula:4}
  \text{V}^{(1)}_p = \text{MultiHead}(\textbf{I}(P),\textbf{I}(P),\textbf{I}(P)),\\
  \text{O}^{(1)}_p = \text{FFN}(\text{V}^{(1)}_p),\\
  \text{FFN}(x) = max(0, xW_1+b_1)W_2+b_2,
\end{gather}
where $\text{V}^{(1)}$ is the first layer result of the multi-head self-attention and $\text {I}(\cdot)$ is the embedding function of the input. The input embedding for word $w_i$ is the sum of its word embedding and position embedding. $\text{O}^{(1)}$ denotes the output of the first layer feed-forward network.
For other layers:
\begin{gather}
\label{formula:7}
  \text{V}^{(n)}_p = \text{MultiHead}(\text{O}^{(n-1)}_p),\text{O}^{(n-1)}_p),\text{O}^{(n-1)}_p),\\
  \text{O}^{(n)}_p = \text{FFN}(\text{V}^{(n)}_p),
\end{gather}
where $\text{n}=$2,...,$N_G$.
We applied layer normalization to  each sublayer by $\text{LayerNorm}(x+\text{Sublayer}(x))$. $Q$ is encoded in the same way.
After $N_G$ identical layers, we can get the final representations $\text{O}^{(N_G)}_p$ and $\text{O}^{(N_G)}_q$, where $\text{O}^{(N_G)}_p$ and $\text{O}^{(N_G)}_q$ are the encoded persona and encoded query, respectively.

\subsubsection*{Tuple-Interaction Decoder}
\label{sec:TLAD}
In the decoding phase, there are three types of information, persona $P$, query $Q$, and response $Y$, which make up a tuple ($P$,$Q$,$Y$). Accordingly, three inter-sentence relations need to be considered: (1) The alignment between $Q$ and $Y$ is beneficial to yield better results~\citep{bahdanau2014attention}. (2) As the persona is composed of several sentences and describes different aspects, we need to find the most relevant persona information according to the relations between P and Y. (3) We also want to know whether the query needs to be answered with the given persona. Thus we should take the relations between $P$ and $Q$ into account.

Considering the above factors, we design a two-layer tuple-interaction mechanism in the decoder, as shown in the first part of Figure~\ref{fig:overall_model}. There are three attentions in two layers: query attention ($\text {Q-Attn}$) and persona attention ($\text {P-Attn}$) in the first layer, and persona-query attention ($\text {PQ-Attn}$) in the second layer. $N_G$ such identical layers compose of the decoder. For the first layer:
\begin{gather}
\label{formula:9}
  \text{V}^{(1)}_{y} = \text{MultiHead}(\textbf{I}(Y),\textbf{I}(Y),\textbf{I}(Y)),\\
  \text{E}^{(1)} = \text{MultiHead}(\text{V}^{(1)}_y,\text{O}^{(N_G)}_p,\text{O}^{(N_G)}_p),\\
  \text{F}^{(1)} = \text{MultiHead}(\text{V}^{(1)}_y,\text{O}^{(N_G)}_q,\text{O}^{(N_G)}_q),\\
  \text{T}^{(1)} = \text{MultiHead}(\text{E}^{(1)},\text{F}^{(1)},\text{F}^{(1)}),\\
  \text{O}^{(1)}_{dec} = \text{FNN}(\text{mean}(\text{E}^{(1)},\text{F}^{(1)},\text{T}^{(1)})),
\end{gather}
where $\text{E}^{(1)}$ and $\text{F}^{(1)}$ are the results of the first layer $\text{P-Attn}$ and $\text{Q-Attn}$. $\text{T}^{(1)}$ is the result of the first layer $\text{PQ-Attn}$. $\text{O}^{(1)}_{dec}$ denotes the first layer output. Note that the $Y$ here is masked to ensure depending only on the known words~\citep{vaswani2017attention}. Repeatedly, for other layers:
\begin{gather}
\label{formula:14}
  \text{V}^{(n)}_{y} = \text{MultiHead}(\text{O}^{(n-1)}_{dec}),\text{O}^{(n-1)}_{dec}),\text{O}^{(n-1)}_{dec}),\\
  \text{O}^{(n)}_{dec} = \text{FNN}(\text{mean}(\text{E}^{(n)},\text{F}^{(n)},\text{T}^{(n)})),
\end{gather}
where $\text{n}=$2,...,$N_G$. After $N_G$ layers, the decoder output $\text{O}^{(N_G)}_{dec}$ is projected from hidden size to vocabulary size, then followed up by a $\text{softmax}$ function to get the words' probabilities:
\begin{gather}
\label{formula:16}
  {\text{Prob}}^{(1)}= \text{SoftMax}(\text{O}^{(N_G)}_{dec}W_3+b_3),
\end{gather}
where $W_3$ is a $\text{hidden size} \times \text{vocabulary size}$ weight matrix and $b_3$ is the bias term with $\text{vocabulary size}$ dimension. And $\text{Prob}^{(1)}$ denotes the output distribution of the first stage. Now we can get the response prototype $\hat{Y}^{(1)}$ from the $\text{Prob}^{(1)}$.

\begin{figure}
\centering
\includegraphics[width=.96\columnwidth]{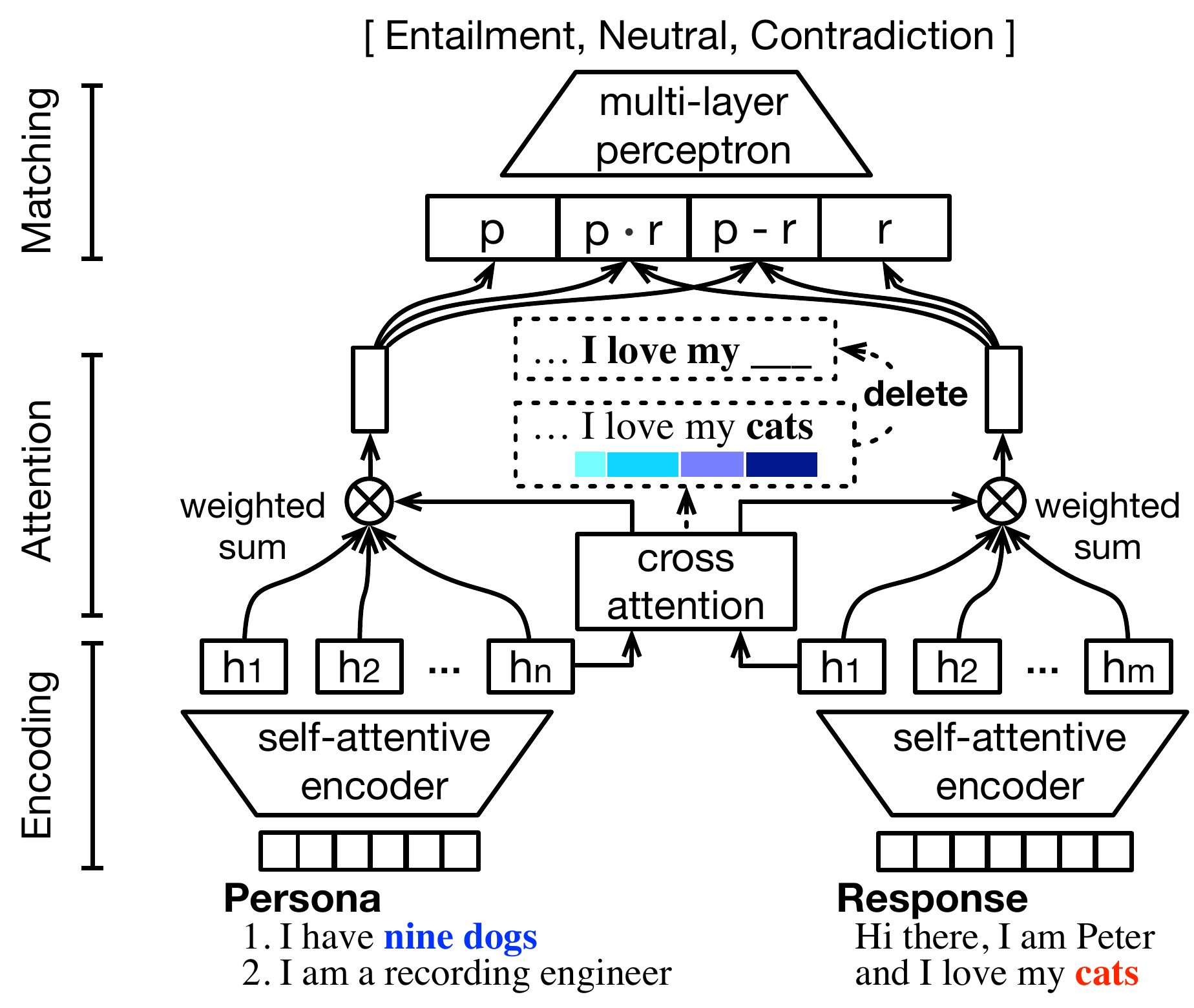}
\caption{ The architecture of our consistency matching model. ``$\cdot$'' and ``$-$'' denote element-wise product and  difference. The dotted line shows inference process, including consistency matching and word deleting.
}
\label{fig:matching_model}
\end{figure}

\subsection{Delete: Consistency Matching Model}
The goal of the consistency matching model $\text{D}$ is to reveal word-level consistency between the persona texts and the response prototype, thus the inappropriate words can be deleted from the prototype. 

This model is trained to estimate the sentence-level entailment category~\citep{bowman2015large} of a response for the given persona texts, which includes {\it entailment}, {\it neutral} and {\it contradiction}. The key is that if the category is not {\it entailment}, we can delete the most contributing words by replacing them with a special mask token, thus giving the model a chance to rephrase. The attention weights can measure each word's contribution.

The architecture of our consistency matching model is shown in Figure~\ref{fig:matching_model}. From bottom to top are the self-attentive encoding layer, cross attention layer, and consistency matching layer.

As described in section~\ref{sec:SAE}, the self-attentive encoder ($\text{SAE}(\cdot)$) performs self-attention over the input to get sentence representations. Because the task of consistency matching is quite different from dialogue generation, we did not share the encoders between the generator $\text G$ and matching model $D$:
\begin{gather}
\label{formula:17}
  \bar{\text{A}}= \text{SAE}_D(P),\\
  \bar{\text{B}}= \text{SAE}_D(\hat{Y}^{(1)}),
\end{gather}
where $\bar{\text{A}}$ is a $\text{hidden size} \times \text{n}$ matrix. $\bar{\text{A}}=[\bar a_1,\bar a_2,...,\bar a_n]$ and $\bar{\text{B}}=[\bar b_1,\bar b_2,...,\bar b_m]$. The $n$ and $m$ are the number of words in persona $P$ and response prototype $\hat{Y}^{(1)}$. Here we applied average pooling stragety~\citep{liu2016learning,chen2017enhanced} to get the summary representations:
\begin{gather}
\label{formula:19}
  \bar{\text{a}}_{0}= \sum_{i=1}^{n}\frac{\bar{a}_i}{n},
\end{gather}
and we can get the response attention weights and attentive response representations by:
\begin{gather}
\label{formula:21}
  {\text{W}_b}= \bar{\text{a}}_{0}^\top \bar{\text{B}},\\
  \widetilde{\text{B}}= {\text{W}_b}\bar{\text{B}}^\top,
\end{gather}
where ${\text{W}_b}$ is attention weights and $\widetilde{\text{B}}$ is response representations. 
Similarly, we can get ${\text{W}_a}$ and $\widetilde{\text{A}}$.

Once $\widetilde{\text{A}}$ and $\widetilde{\text{B}}$ are generated, three matching methods~\citep{chen2017enhanced} are applied to extract relations: concatenation, element-wise product, element-wise difference. The results of these matching methods are concatenated to feed into a  multi-layer perceptron, which has three layers and tanh activation in between. The output is followed up by a SoftMax function to produce probabilities.

In the inference process, as shown in Figure~\ref{fig:matching_model}, the response attention weights $\text{W}_b$ is leveraged to illustrate the inconsistent words, which will be deleted\footnote{In this paper, ``delete'' a word means replacing this word with a special mask token.}. In practice, we use a simple heuristic rule for deleting words: as long as the category is not $entailment$, we will delete 10\% of the words (at least one word)\footnote{In our experiments, we found that deleting more words made it difficult for rewriter R to learn.}, with the highest attention weight, in the prototype $\hat{Y}^{(1)}$. In this way, we get the masked prototype $\hat{Y}^{(2)}$.

\subsection{Rewrite: Masked Prototype Rewriter}
The rewriter $\text{R}$ takes the masked prototype and persona texts as input and outputs the final response.

$\text{R}$ is also a transformer decoder, which is similar to the decoder of ${\text G}$ in section~\ref{sec:TLAD}, but with a minor difference: the masked prototype is close to the target response, thus the direct attention between the prototype and target response is needless. The architecture of ${\text R}$ can be seen in the third part of Figure~\ref{fig:overall_model}, which can be formalized as:
\begin{gather}
\label{formula:22}
  \text{O}^{(N_G)}_{mp} = \text{SAE}_G(\hat{Y}^{(2)}),\\
  \text{V}^{(n)} = \text{MultiHead}(\text{O}^{(n-1)}_{rw}),\text{O}^{(n-1)}_{rw}),\text{O}^{(n-1)}_{rw}),\\
  \text{S}^{(n)} = \text{MultiHead}(\text{V}^{(n)},\text{O}^{(N_G)}_p,\text{O}^{(N_G)}_p),\\
  \text{K}^{(n)} = \text{MultiHead}(\text{S}^{(n)},\text{O}^{(N_G)}_{mp},\text{O}^{(N_G)}_{mp}),\\
  \text{O}^{(n)}_{rw} = \text{FNN}(\text{mean}(\text{S}^{(n)},\text{K}^{(n)})),
\end{gather}
where $\text{O}^{(N_G)}_{mp}$ is the encoded masked prototype and $\text{SAE}_G$ is the self-attentive encoder of $\text G$. $\text{O}^{(N_G)}_p$ is the encoded persona. After $N_R$ identical layers, the same generation process as in $\text G$ is applied to the $\text{O}^{(N_R)}_{rw}$, and we can get the final response $\hat{Y}^{(3)}$.

\subsection{Training}
The consistency matching model $\text D$ is trained separately from the prototype generator $\text G$ and rewriter $\text R$. As forementioned, the matching model $\text D$ is trained in a natural language inference fashion on the DNLI dataset~\citep{WelleckDNLI}, which has been well defined by the previous studies~\citep{bowman2015large,chen2017enhanced,gong2018DIIN}. We minimize the CrossEntropy loss between the outputs of $\text D$ and the ground truth labels.

The $\text G$ and $\text R$ share the same training targets. We trained them by the standard maximum likelihood estimate.
Notice that there are two different deleting strategies in training: (1) In the warm-up phase, because the $\text G$ can hardly generate high-quality prototypes at this period, we randomly delete each word, with a 10\% probability, from the prototype. (2) After that, the trained consistency matching model $\text D$ is leveraged to delete words.

\begin{table}[t]
\centering
\begin{tabular}{l|lll}
\toprule
{\bf Data}&{\bf Train}&{\bf Valid}&{\bf Test}\\
\midrule
Persona Texts 	& 74,522  & 5,843  & 4,483 \\
Q-R Pairs  		& 121,880 & 9,558  & 7,801 \\
\bottomrule
\end{tabular}
\caption{Some statistics of Persona-Chat dataset. Valid denotes Validate and Q-R denotes Query-Response.}
\label{tab:dataset_personachat}
\end{table}

\begin{table}[t]
\centering
\begin{tabular}{l|lll}
\toprule
{\bf Label}&{\bf Train}&{\bf Valid}&{\bf Test}\\
\midrule
Entailment      & 100,000 & 5,500 & 5,400 \\
Neutral  		& 100,000 & 5,500 & 5,400 \\
Contradiction 	& 110,110 & 5,500 & 5,700 \\
\bottomrule
\end{tabular}
\caption{Key statistics of DNLI dataset.}
\label{tab:dataset_dnli}
\end{table}

\section{Experiments}
\subsection{Datasets}

We carried out the persona-based dialogue generation experiments on the public available Persona-Chat dataset~\citep{zhang-2018-persona}. Furthermore, we trained the consistency matching model on the recently released Dialogue Natural Language Inference (DNLI) dataset~\citep{WelleckDNLI}.

We show the statistics of the Persona-Chat dataset in Table~\ref{tab:dataset_personachat}.
The DNLI dataset~\citep{WelleckDNLI} is an enhancement to the Persona-Chat. It is composed of {\it persona-utterance} pairs from the Persona-Chat, and these pairs are further labeled as {\it entailment}, {\it neutral}, and {\it contradiction}. Some statistics of this dataset are given in Table~\ref{tab:dataset_dnli}.

\subsection{Compared Models}
To the best of our knowledge, this is an early work in modeling explicit persona consistency. To show the effectiveness of our models, we mainly compare it with the persona-based dialogue models:
\begin{itemize}
  \item {{\bf S2SA}} S2SA is an RNN-based attentive seq2seq model~\citep{bahdanau2014attention}. It only takes the query as input.
  \item {{\bf Per-S2SA}} This is a seq2seq model that prepends all persona texts to the query as input~\citep{zhang-2018-persona}.
  \item {{\bf GPMN}} Generative Profile Memory Network is an RNN-based model that encodes persona texts as individual memory representations in a memory network~\citep{zhang-2018-persona}.
  \item {{\bf DeepCopy}} An RNN-based hierarchical pointer network, which leverages copy mechanism to integrate persona~\citep{yavuz2019deepcopy}.
  \item {{\bf Per-CVAE}} This is a memory augmented CVAE model to exploit persona texts for diverse response generation~\citep{song-percvae}.
  \item {{\bf Transformer}} Different from the RNN-based models, transformer is a self-attention based sequence transduction model~\citep{vaswani2017attention}. The persona texts are concatenated to the query to serve as its input.
\end{itemize}

\subsection{Experimental Settings}
For all the RNN-based baseline models, they are implemented by two-layer LSTM networks with a hidden size 512. 
For the Transformer, the hidden size is also set to 512, and the layers of both encoder and decoder are 3. The number of heads in multi-head attention is 8, and the inner-layer size of the feedforward network is 2048.
The word embeddings are randomly initialized, and the embedding dimension of all models is set to 512.

Our model applies the same parameter settings as the transformer. The number of layers $\text N_G=\text N_D=\text N_R=3$. G and R share the word embeddings, but the matching model D uses independent embeddings.
We use token-level batching with a size 4096.
Adam is used for optimization, and the warm-up steps are set to 10,000.
We implemented the model in {\it OpenNMT-py}~\citep{klein-etal-2017-opennmt}.

\subsection{Evaluation Metrics}

In the evaluation, there are two essential factors to consider: {\bf persona consistency} and {\bf response quality}. We apply both human evaluations and automatic metrics on these two aspects to compare different models.

\begin{table*}[t]
\centering
\begin{tabular}{l|llll|lllll}
\toprule
{\bf Model}&{\bf Const.}&{\bf Fluc.}&{\bf Relv.}&{\bf Info.}&{\bf PPL}&{\bf Dist-1.}&{\bf Dist-2.}&{\bf $\text{Ent}_\text{diin}$}&{\bf $\text{Ent}_\text{bert}$}\\
\midrule
S2SA             & 15.9\% & 3.17 & 2.84 & 2.63 & 34.8 & 1.92 & 4.86 & 9.80\% & 1.83\% \\
GPMN             & 34.8\% & 3.78 & 3.57 & 3.76$^\dagger$ & 34.1 & 1.89 & 7.53 & 14.5\% & 7.36\%   \\
Per-S2S          & 35.3\% & 3.43 & 3.22 & 3.32 & 36.1 & 2.01 & 7.31 & 13.5\% & 6.15\%  \\
DeepCopy         & 36.0\% & 3.26 & 3.08 & 2.87 & 41.2 & 2.35 & 8.93 & 16.7\% & 8.81\% \\
Transformer      & 38.8\% & 3.46 & 3.65$^\dagger$ & 3.54 & 27.9 & 3.12 & 15.8 & 14.2\% & 9.52\% \\
Per-CVAE         & 42.7\% & 3.53 & 2.97 & 3.66 & -$^*$ & {\bf3.83}$^\dagger$ & 20.9 & 17.2\% & 7.36\% \\
\midrule
{GDR (ours)}     & {\bf49.2\%} & {\bf3.86} & {\bf3.68} & {\bf3.77} & \bf{16.7} & 3.66 & {\bf22.7} & {\bf21.5\%} & {\bf13.0\%}  \\
\bottomrule
\end{tabular}
\caption{Results of human evaluations (on the left) and automatic metrics (on the right). The Dist-1.\& 2. are scaled by $10^{-2}$. Significant tests (t-test) are performed, and our method is significantly better than all methods on most metrics (p-value$<$0.05), with the exceptions marked by $\dagger$. We also present two model-based ratios, the $\text{Ent}_\text{diin}$ and the $\text{Ent}_\text{bert}$, as an additional reference for persona consistency assessments. Note that the automatic metrics are calculated on the whole test set.
{\bf*} The sampling process in CVAE leads to very unstable PPL.}
\label{tab:main}
\end{table*}

\paragraph{Human Evaluation} We recruit five professional annotators from a third-party company. These annotators have high-level language skills but know nothing about the models. We sampled 200 {\it persona-query-response} tuples per model for evaluation. Duplicated queries (such as greetings which appear more than once) will not be sampled twice.

First, we evaluate the persona consistency of a response. The annotators are asked to decide whether the response is consistent with the given persona. 0 indicates irrelevant or contradictory and 1 indicates {\bf consistent} (Const.).

Second, we evaluate the quality of a response on three conventional criteria: {\bf fluency} (Fluc.), {\bf relevance} (Relv.), and {\bf informativeness} (Info.). Each aspect is rated on five-scale, where 1, 3, and 5 indicate unacceptable, moderate, and excellent performance, respectively. 2 and 4 are used for unsure.

\paragraph{Automatic Metrics} \citet{dziri2019evaluating} has shown that natural language inference based {\it entailment} ratio can be used as an indicator of dialogue consistency. Here we trained two well-performed NLI models, DIIN~\citep{gong2018DIIN} and BERT~\citep{devlin2019bert}, to automatically predict the category of {\it persona-response} pairs, and we calculated the ratio of {\it entailment} as an additional reference to the persona consistency. In our experiments, DIIN and BERT achieved 88.78\% and 89.19\% accuracy on the DNLI test set, respectively, compared with previous best results 88.20\%. 
The trained models are then leveraged for calculating {\it entailment} ratios. Two model-based {\it entailment} ratios are abbreviated as {\bf $\text{Ent}_\text{diin}$} and {\bf $\text{Ent}_\text{bert}$}.

For dialogue quality, we follow~\citet{zhang-2018-persona} to use {\bf perplexity} (PPL) to measure the fluency of responses. Lower perplexity means better fluency. Besides, we also use {\bf Dist-1} / {\bf Dist-2}~\citep{li2016diversity} to examine the model's ability to generate diverse responses, which is the ratio of distinct uni-grams / bi-grams.

\begin{table}[t]
\centering
\begin{tabular}{l|lll}
\toprule
{{\bf GDR} vs}&{\bf Win(\%)}&{\bf Tie(\%)}&{\bf Lose(\%)}\\
\midrule
S2SA        & 48.0 & 38.2 & 13.8 \\
Per-CVAE    & 46.1 & 29.8 & 24.1 \\
DeepCopy    & 43.8 & 35.5 & 20.7 \\
Per-S2S     & 41.3 & 36.1 & 22.6 \\
GPMN        & 35.0 & 31.0 & 34.0 \\
Transformer & 34.7 & 32.1 & 33.2 \\

\bottomrule
\end{tabular}
\caption{GDR response quality gains over other baseline methods on a pairwise human judgment.}
\label{tab:pairwise}
\end{table}

\subsection{Main Results}
We report the main evaluation results in Table~\ref{tab:main}. Compared with baseline methods, our GDR model obtains the highest consistency score of 49.2\% in human evaluation, which is significantly better than other methods. The target responses in the sampled data are also annotated, and 65.4\% of them expressed persona information. Moreover, the two model-based entailment ratios, which are calculated on the whole test set, also prove the effectiveness of our GDR model. Although the two NLI models differ in results, our GDR model ranks first under the evaluation of both DIIN and BERT.

For dialogue quality, our proposed model has a remarkably lower perplexity of 16.7 than all other baseline methods. An analysis can be seen in Section~\ref{sec:analysis}. Besides, our distinct-2 metric is even significantly better than the Per-CVAE model, which is designed to generate diverse responses.

Additionally, we carried out pairwise response comparison to see the dialogue quality gains. We report the results in Table~\ref{tab:pairwise}. While the GDR model significantly improves persona consistency, it can still generate high-quality responses like the transformer and GPMN.

\subsection{More Analysis}
\label{sec:analysis}
As the proposed model achieves better performance than baseline methods, we turn to ablation tests to further quantify the contributions made by different components. We ablated our model through several different approaches:
\begin{itemize}
  \item {{\bf GR}} It removes the matching model D, i.e., generates a prototype and rewrites it directly.
  \item {{\bf GRdR}} This approach replaces the matching model D with 10\% random deleting (Rd), thus to see if the masked prototype, extracted by our matching model D, is beneficial.
  \item {{\bf G}} Our model's generator, without further consistency matching and rewriting.
  \item {{\bf T}} It is a transformer generator but removes the tuple-interaction in section~\ref{sec:TLAD} and directly concatenates persona texts to the query. This model is equivalent to the vanilla transformer.
\end{itemize}

\begin{table}[t]
\centering
\begin{tabular}{l|lllll}
\toprule
{\bf Model}&{\bf Const.}&{\bf Fluc.}&{\bf Relv.}&{\bf Info.}&{\bf PPL}\\
\midrule
GDR & 49.2\% & 3.86 & 3.68 & 3.77 & 16.7\\
\midrule
GR & 42.4\% & 3.72 & 3.40 & 3.66 & 18.0\\
GRdR & 40.0\% & 3.60 & 3.29 & 3.56 & 20.6\\
G & 40.1\% & 3.69 & 3.35 & 3.55 & 26.3\\
T & 38.8\% & 3.46 & 3.65$^\ddagger$ & 3.54 & 27.9\\
\bottomrule
\end{tabular}
\caption{Results of the ablation study. GDR is significantly better than the ablated approaches, with an only exception marked by $\ddagger$.}
\label{tab:ablation_human}
\end{table}

\begin{table}[t]
\centering
\begin{tabular}{l|lll}
\toprule
{{\bf GDR} vs}&{\bf Win(\%)}&{\bf Tie(\%)}&{\bf Lose(\%)}\\
\midrule
GRdR  & 41.7 & 39.5 & 18.8 \\
GR    & 39.9 & 40.9 & 19.2 \\
G     & 38.1 & 35.8 & 26.1 \\
\bottomrule
\end{tabular}
\caption{Pairwise human judgment on response quality.}
\label{tab:pairwise_ablated}
\end{table}

We report the results in Table~\ref{tab:ablation_human}. First, we look into which components contribute to the consistency. As seen, from T, G, GR to GDR, every step has an observable improvement in {\it Const.}, indicating the effectiveness of our model's design. Both the tuple-interaction in G and the rewriting process in R contribute to the improvements of persona consistency. The GRdR approach, with nothing different from GDR but a random deleting strategy, serves as a foil to our GDR model, which indicates a well-learned consistency matching model is of great benefit to our three-stage generation framework to generate persona consistent dialogues.

Second, we investigated the improvement of our perplexity. As we can see, the one-stage transformer approaches G and T have a perplexity higher than 26. In contrast, after we add the rewriter R, the perplexity of all approaches has a significant decline, no matter whether there is a matching model D. Lower perplexity means lower cross-entropy, which indicates the responses from the models have more ground truth words. To some extent, perplexity verifies the human evaluation results of the consistency. 
One reason for this improvement could be that the rewriter works like a denoising autoencoder~\citep{vincent2008denoising}, and it can focus more on the reconstruction of the missing information of sequence itself, rather than learning to map a sequence to an entirely different one.

We observed that the relevance scores of GR, GRdR, and G are a little inferior to the T. Even the GDR model is not significantly better than T on the relevance score. One plausible explanation is that all these models are specially designed for integrating persona information, although they obtain much better consistency score, it may come at the cost of relevance score.

Moreover, we compared the GDR's response quality with three ablated models and reported it in Table~\ref{tab:pairwise_ablated}. As we can see, the deleting and rewriting, which are designed for improving consistency, also have a positive effect on the dialogue quality.

\begin{table}[t]
\small
\centering
\begin{tabular}{r|l}
\toprule
{\bf Persona}&{i. \ My mother is a dentist}\\
{}&{ii. I'm currently looking for a job}\\
{\bf Query}&{\bf I want to become a physical therapist.}\\
\midrule
{Gen\&Del}&{\colorbox[rgb]{0.9921,0.9608,0.9020}{I}\colorbox[rgb]{1.0,0.9373,0.8353}{wish}\colorbox[rgb]{0.9921,0.9608,0.9020}{I}\colorbox[rgb]{0.9921,0.9608,0.9020}{could}\colorbox[rgb]{1.0,0.9373,0.8353}{be}\colorbox[rgb]{0.9921,0.9608,0.9020}{a}\colorbox[rgb]{1.0,0.8078,0.6235}{\st{nurse}}\colorbox[rgb]{0.9921,0.9608,0.9020}{.}}\\
{Rewrite}&{I want to be \underline{a dentist} \underline{like my mom}.}\\
\midrule
{w/o Delete}&{I want to become a nurse like you.}\\
\midrule
Transfmr   & I hear you. I am a nurse and a mom.\\
Per-S2S       & What kind of business? \\
GPMN          & I am a professional painter, what do you do? \\
DeepCopy      & What do you do for a living? \\
Per-CVAE      & I guess I would buy you one \\
\midrule
\midrule
{\bf Persona}&{i. \ Gymnastics is my favorite sport}\\
{}&{ii. I'm a gold medalist olympian}\\
{\bf Query}&{\bf I am a mini van driver and I drive around.}\\
\midrule
{Gen\&Del}&{\colorbox[rgb]{0.9921,0.9608,0.9020}{That's}\colorbox[rgb]{0.9921,0.9608,0.9020}{awesome}\colorbox[rgb]{0.9921,0.9608,0.9020}{.}\colorbox[rgb]{1.0,0.9373,0.8353}{I}\colorbox[rgb]{0.9921,0.9608,0.9020}{also}\colorbox[rgb]{1.0,0.8941,0.7686}{love}\colorbox[rgb]{0.9921,0.9608,0.9020}{to}\colorbox[rgb]{1.0,0.8078,0.6235}{\st{ski}}\colorbox[rgb]{0.9921,0.9608,0.9020}{.}}\\
{Rewrite}&{That's cool. I am a \underline{champion} \underline{gymnast}.}\\
\midrule
{w/o Delete}&{That's awesome. I am going to ski.}\\
\midrule
Transfmr & Oh. I am still in high school . \\
Per-S2S     & I drive a big van, but around the house. \\
GPMN        & I drive a FIAT. I love it. \\
DeepCopy    & What do you do for a living? \\
Per-CVAE    & That is why you like dogs? Tell me.\\
\midrule
\midrule
{\bf Persona}&{i. \ I like to grill outdoors}\\
{}&{ii. I have been retired for years}\\
{\bf Query}&{\bf Old enough to work in construction. you?}\\
\midrule
{Gen\&Del}&{\colorbox[rgb]{0.9921,0.9608,0.9020}{I'm}\colorbox[rgb]{1.0,0.8078,0.6235}{retired}\colorbox[rgb]{0.9921,0.9608,0.9020}{.}\colorbox[rgb]{0.9921,0.9608,0.9020}{I}\colorbox[rgb]{1.0,0.8941,0.7686}{spend}\colorbox[rgb]{0.9921,0.9608,0.9020}{time}\colorbox[rgb]{1.0,0.8078,0.6235}{outdoors}\colorbox[rgb]{0.9921,0.9608,0.9020}{.}}\\
{Rewrite}&{I'm \underline{retired}. I spend most of my time \underline{outdoors}.}\\
\midrule
{w/o Delete}&{I'm retired. I spend most of my time outdoors.}\\
\midrule
Transfmr  & I am a boy and work in the construction. you? \\
Per-S2S     & No, I am a student. \\
GPMN        & I am a teacher, I love to be a teacher. \\
DeepCopy    & I work in the construction industry. \\
Per-CVAE    & I am a retired officer I love my bike ride.\\
\bottomrule
\end{tabular}
\caption{Example responses from different models, with a visualization of the consistency matching weights. Strikethrough words are the masked words in Delete stage. The {\it w/o Delete} is the ablated model GR in section~\ref{sec:analysis}, and {\it Transfmr} is short for Transformer.
}
\label{tab:cases}
\end{table}

At last, we presented some generated examples in Table~\ref{tab:cases}, together with the visualization of attention weights from match module D.
In the first case, although the generated prototype is {\it neutral} regarding the persona, the word ``nurse'' is still masked according to our strategy. And after the rewriting stage, the final response expresses persona. In the second case, the prototype is potentially contradictory to the persona, and the keyword is successfully deleted by the matching model D.
In the third case, the prototype is consistent with the persona, and no word is deleted. As a result, the final output response is the same as the output of no deletion model GR.
In these cases, both consistency and quality are improved after the final rewriting.

\section{Conclusion and Future Work}

In this paper, we presented a three-stage framework, Generate-Delete-Rewrite, for persona consistent dialogue generation. Our method adopts transformer architecture and integrates a matching model to delete the inconsistent words. Experiments are carried out on public-available datasets. Both human evaluations and automatic metrics show that our method achieves remarkably good performance. In the future, we plan to extend our approach to improve the consistency of multi-turn dialogues.

\section*{Acknowledgments}

This paper is supported by the National Natural Science Foundation of China under Grant No.61772153 and No.61936010. Besides, we want to acknowledge the Heilongjiang Province Art Planning Project 2019C027 and the Heilongjiang Province Social Science Research Project 18TQB100.
We also would like to thank all the anonymous reviewers for their helpful comments and suggestions.

\bibliography{acl2020}
\bibliographystyle{acl_natbib}

\end{document}